# Introducing the Large Medical Model:

State of the art healthcare cost and risk prediction
with transformers trained on patient event sequences

GenHealth.ai

Ricky Sahu, Eric Marriott, Ethan Siegel, David Wagner, Flore Uzan, Troy Yang, Asim Javed
{hello} @genhealth.ai

## Abstract

With U.S. healthcare spending approaching $5T ("NHE Fact Sheet" 2024), and 25% of it estimated to be wasteful ("Waste in the US the health care system: estimated costs and potential for savings", n.d.), the need to better predict risk and optimal patient care is evermore important. This paper introduces the Large Medical Model (LMM), a generative pre-trained transformer (GPT) designed to guide and predict the broad facets of patient care and healthcare administration. The model is trained on medical event sequences from over 140M longitudinal patient claims records with a specialized vocabulary built from medical terminology systems and demonstrates a superior capability to forecast healthcare costs and identify potential risk factors. Through experimentation and validation, we showcase the LMM's proficiency in not only in cost and risk predictions, but also in discerning intricate patterns within complex medical conditions and an ability to identify novel relationships in patient care. The LMM is able to improve both cost prediction by 14.1% over the best commercial models and chronic conditions prediction by 1.9% over the best transformer models in research predicting a broad set of conditions. The LMM is a substantial advancement in healthcare analytics, offering the potential to significantly enhance risk assessment, cost management, and personalized medicine.

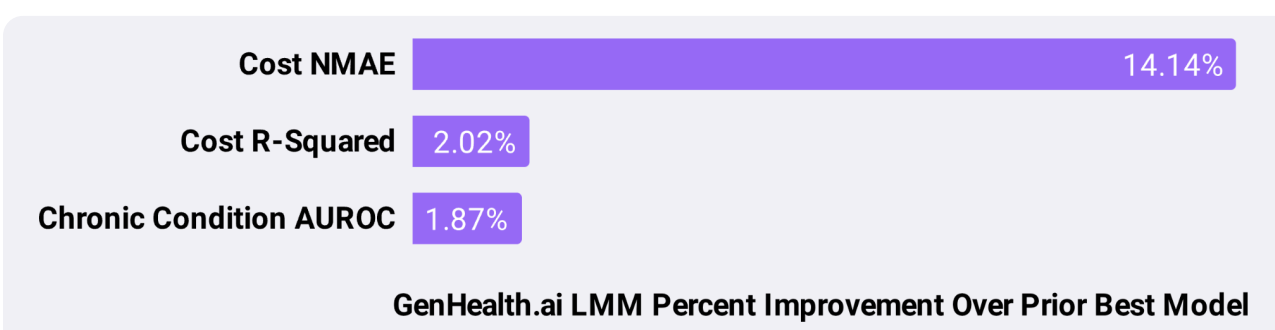

**Result summary for Large Medical Model evaluation metrics**

## 1. Introduction

Healthcare administration involves balancing the cost of care with the need to mitigate risks and ensure positive patient outcomes . The industry, especially in the United States, is largely driven by health plans attempting to reduce healthcare costs while maintaining high margins, and healthcare providers (typically under fee-for-service incentive structures) who benefit from increased services and care resulting in increased revenue. Pharmaceutical companies (pharma) are also deeply invested in this landscape, aiming to develop and commercialize medications to improve patient outcomes while mitigating adverse reactions. New value-based-care incentives propagate the balancing act between cost and risk, but shift the responsibility to single entities like Accountable Care Organizations (ACOs) or Medicare Advantage Plans.

For providers, payers, pharma, and patients to better balance risk, cost, and treatment efficacy, they must understand what attributes of an individual's history most impact the future, and the web of interaction effects among those over time. Along with medical outcomes, the financial aspects of care delivery are also important. They are, however, often separately codified in contracts between payers, providers, and pharma, further occluded by negotiations between the parties. Understanding the long term cause and effect of healthcare interventions is an exceptionally difficult problem for humans and machines alike.

First, the science of medicine is continuously developing. Our understanding of how different parts of the body react to various treatments and interventions continues to evolve. As new therapies and medications are developed, the complexity of medical care increases, necessitating ongoing learning and adaptation. Healthcare's dynamic environment makes it challenging to maintain up-to-date knowledge while considering a broad set of possibilities and effective application of treatments across diverse patient populations.

Second, the data available in healthcare is vast and varied. Medical records encompass an extensive range of data types, including procedures, medications, costs, locations, provider information, genomics, conditions, laboratory results, textual notes, demographics, images, and more. Furthermore, these data originate from multiple sources such as electronic medical records (EMRs), insurance claims, sensors, and the environment. As longitudinal patient records are often fragmented it is difficult to assemble a comprehensive view of a patient's medical history and predict future healthcare needs accurately.

Finally, the variation in the population and environment leads to highly individualized medical journeys that even large-scale research studies fail to capture fully.The research that is done is often isolated geographically or to only those patients who will respond most favorably, limiting the generalizability of findings. This individuality necessitates personalized approaches to care that account for the distinct medical and social contexts of each patient.

To achieve improved healthcare predictions within the high dimensional nature of healthcare, one approach is to imagine healthcare data as a sequence of events playing out over time. This framing lends itself naturally to sequence to sequence models in order to predict future events. The field of sequence based modeling has recently been taken over by deep learning and is now largely centered upon sequence to sequence neural network architectures, such as transformers.

## 2. Related Work

Healthcare plays a massive role in national economies, is a major focus of financial investment globally, and is ever important in our own lives as we age. As healthcare costs continue to mount, so has interest from government agencies, insurance providers, healthcare institutions, medical professionals, the pharmaceutical industry, and patients in understanding how to better manage and control these expenses while achieving better outcomes. This imperative has produced many commercial models designed to accurately forecast healthcare costs and patient risks.

Traditionally, healthcare analytics are based on purpose-built statistical models specialized for specific problems. These models typically focus on structured data such as diagnostic codes, demographic information and pharmacy records. For example, the Hierarchical Condition Category (HCC) model developed by the Centers for Medicare and Medicaid Services (CMS) is a classical linear model used for risk adjustment to identify patients with chronic health conditions and estimate their healthcare costs. Since 2004, CMS has used the HCC model to calculate payments to healthcare organizations for patients with Medicare Advantage, Accountable Care Organization (ACO), and Affordable Care Act (ACA) plans ("IMO Health", n.d.). Similarly, models such as the Chronic Illness & Disability Payment System (CDPS), Milliman's MARA, and Adjusted Clinical Groups (ACG) from Johns Hopkins University assess morbidity burdens by analyzing diagnostic data and a limited set of other events like medications and past cost. These models are often designed for targeted populations such as Medicaid beneficiaries and are effective within their limited scope.

While most commercial offerings are also based on linear models, some employ more advanced statistical methods such as gradient boosted trees (Kumar 2023). However, these models still fall short in capturing the complexity, sequencing, and nuances of medical data as they fail to account for the entirety of a patient's health record. This limitation becomes evident in predicting outcomes for complex diseases such as diabetes where long term progression pattern plays significant role (An Tran-Duy 2020,).

Given the billions of dollars to be gained with better cost and risk prediction models and the clear benefits on patient outcomes, the industry has begun to explore the application of generative AI to this domain. Recent works such as BEHRT (BERT for Electronic Health Records) (Li, n.d.,), RETAIN (REverse Time AttentioN) (Choi, n.d.) Foresight (Kraljevic, n.d.) and Deepr (Nguyen, n.d.) have shown promising results in using sequential models and attention mechanisms for healthcare predictions. For instance, BEHRT, adapts the transformer architecture to capture the nuances of patient histories, while RETAIN (REverse Time AttentIoN) employs a reverse time two level attention mechanism to prioritize recent clinical events in its prediction, enhancing the interpretability of the model to understand what influences the predictions.

Deepr applies deep learning techniques to model patient sequences for predictive analytics. Additionally, the potential of applying Large Language Models (LLMs) to healthcare is gaining traction, with researchers investigating how these models can be fine-tuned to handle patient data.

The Foresight v2 model is a LLM fine-tuned on MIMIC-III dataset to predict both disorders and disease risk scores for patient timelines using SNOMED codes. The authors expand the tokenizer to add these codes and report improvements over their Foresight v1 model on a test set of 2101 patients, demonstrating how generative AI can be effective in predicting both specific disorders and overall risk of future medical events. However they also describe challenges due to lack of data and reliance on a small dataset, which may fail to capture the diversity of patient populations.

Despite the progress in exploring generative AI and transformer-based approaches, a critical gap remains–the lack of a transformer-based model trained on multi-modal attributes from claims and clinical records, tokenized using medical terminology systems, and trained at scale. In this research we present the LMM, which achieves state-of-the-art (SoTA) performance for cost and risk prediction in healthcare.

## 3. Method

### 3.1 Large Medical Model (LMM)

To conduct our research, we trained an autoregressive transformer based neural network on patient medical events as a unified sequence of data. Rather than employing language and text in the sequence as is done with LLMs, we compose individual patient timelines by combining structured medical event data as sequences of medical event codes and taxonomies. A text based approach would require much longer sequence lengths to convey the same information presented in medical codable concepts which LLM's have been shown to struggle with (Soroush, n.d.). For example, medical concepts such as ICD-10 CM codes (International Classification of Diseases) convey information about patient diagnoses and conditions. Typically in human language a condition could be referred colloquially as "breast cancer," more specifically for healthcare data standards and billing purposes as a "Malignant neoplasm of unspecified site of unspecified female breast" which is also specified as the ICD-10 CM code "C50.919." Using a language based vocabulary, later more specific text requires 13 tokens, whereas in our approach using a medical event based vocabulary we can convey the same information in 1 to 4 tokens.

An LLM tokenizer would use 25 tokens to convey this information:

47 year old female with a malignant
neoplasm of unspecified site of
unspecified female breast on
metformin which cost $12

By contrast, GenHealth.ai's LMM represents the same information with codable concepts instead of words resulting in a shorter, more information dense sequence:

47 year old female with a malignant
neoplasm of unspecified site of
unspecified female breast on
metformin which cost $12

Autoregressive models tend to place more importance on recent events and tokens, and by reducing the total number of tokens we are able to drastically reduce compute requirements and capture a more salient signal for each event. Additionally, training on text would necessitate getting data from the entire internet, and might occlude healthcare specific insights. Furthermore, text models struggle with numeric data, while our approach allows for a custom tokenization for better handling of numerical information. Finally, text models do not have a built in way for representing time series events, this makes tracking patient historical timelines difficult.

### 3.2 Data

Our patient sequences are produced from healthcare claims data from a number of sources in a variety of formats. The total data set covers 140M patients of which 95% are in the training set and 5% split between the validation and test sets. The patient population spans the entire United States geographically and is composed of approximately equally of commercial, Medicare, and Medicaid members including medical events from 2016 to 2022. Patients aged 0 to over 100 are all included. We do not prune or filter the data when training our model.

Each patient's data is formatted into a linearized sequence of event tokens representing the medical history of an individual. There are tokens representing place of service, service codes (CPT-4, HCPCS, ICD-10 PCS), diagnosis codes (ICD-10 CM), drugs (NDC), demographic information, claim costs and the timing of events.

### 3.3 Monte Carlo Simulations

Subsequent to model training, during inference, the LMM produces multiple sequences of events given a patient's history. Each inference run using a patient's history will produce different future events given the non-deterministic nature of our decoding method. This Monte Carlo simulation of futures allows us to estimate the probabilities of events as we aggregate across them.

Consider the example as shown in figure 1; 5 different sequence runs with the same patient's history may produce 5 different futures which can then be stacked and aggregated to produce a probability distribution of the patient's future over time. The individual events from each of these predictions or the compiled simulations can be used to interrogate the data at the event or sequence level. For example, if each green event were one dollar, we can interpret the model to say in runs 1, 3, and 4 the patient's future set of medical events cost $1, run 2 cost $2, and run 5 did not incur any medical cost. Moreover, we can associate successive events with conditional probabilities beginning to indicate causal relationships.

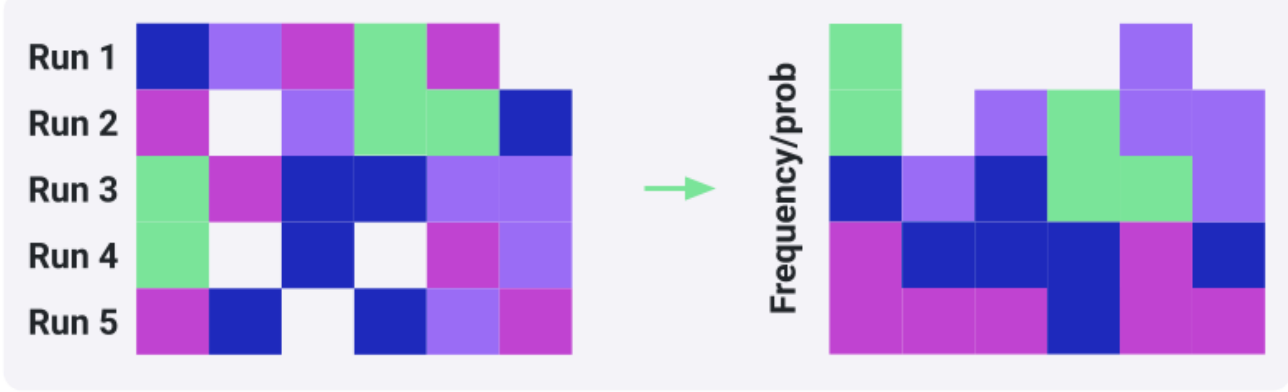

**Figure 1: Monte Carlo Simulation**

In contrast to other models that regress a single variable like cost or produce a probability distribution over a small subset of condition categories, the LMM can generate likelihood of occurrence over any event in the training vocabulary, including drugs, diagnoses, procedures, or other events. When using the model to predict cost, this provides insight into what the cost is for and when it is incurred for any individual or a group of individuals, making the output far more actionable.

### 3.4 Training and Evaluation Pipeline

Overall the process consisted of this series of steps:

1. Build a vocabulary of medical events
2. Construct sequences of historical patient events
3. Train a transformer based neural network model on those events to predict the next token
4. Use the resulting model to run inference multiple times for each patient in a validation set
5. Calculate the total cost of care by summing the predicted cost events, and compare to the actual

## 4 Experiments and Results

The study is divided into two parts: one focused on forecasting healthcare costs and the other on predicting chronic conditions. Each part uses a distinct cohort, chosen to align with the objectives of the respective analysis and directly compare to prior research.

### 4.1 Care Cost Prediction

We benchmarked the LMM against existing predictive risk models as measured by the Society of Actuaries (SoA) in Accuracy of Claims-Based Risk Scoring Models (2016). To the best of our knowledge, no equally comprehensive evaluation of industry standard risk scoring models has been published since the SoA study. Our model outperforms all the models compared in the SoA paper, including thosefrom Milliman, Hopkins, Cotiviti, etc.

We conducted additional analysis comparing the public Hierarchical Condition Category (HCC) risk model from the US government, and found that we outperform that as well. Additionally, given the meaningful advancements in Large Language Models (LLMs) we also evaluated various methods of coaxing Open AI's gpt-4o model into predicting next year's cost given last year's data (in both structured and language formats). However, the best response from the LLM was to use last year's total cost to predict next year's total cost, which is a naive approach that performs far worse than all other models in this study.

#### 4.1.1 Data Collection

To compare to the results from the SoA paper, we produced a cohort of 50,000 patients per the same inclusion and exclusion criteria as specified in the SoA paper, from a similarly broad commercial patient population. While the 2016 SoA study used data from 2012 to predict healthcare costs for 2013, we used the data from 2017 to predict 2018, as our data begins in 2016.

The criteria are as follows:

- Individuals who did not have prescription drug data available in the dataset were excluded.
- Any individual with at least one capitated service in either year of the historical data was excluded.
- Exclude individuals with less than 12 months of enrollment in the baseline year (2017)

By aligning our cohort selection with the criteria used in the SoA study, we ensure that our findings are comparable and that our model's performance can be benchmarked against established standards in the field.

#### 4.1.2 Data Overview

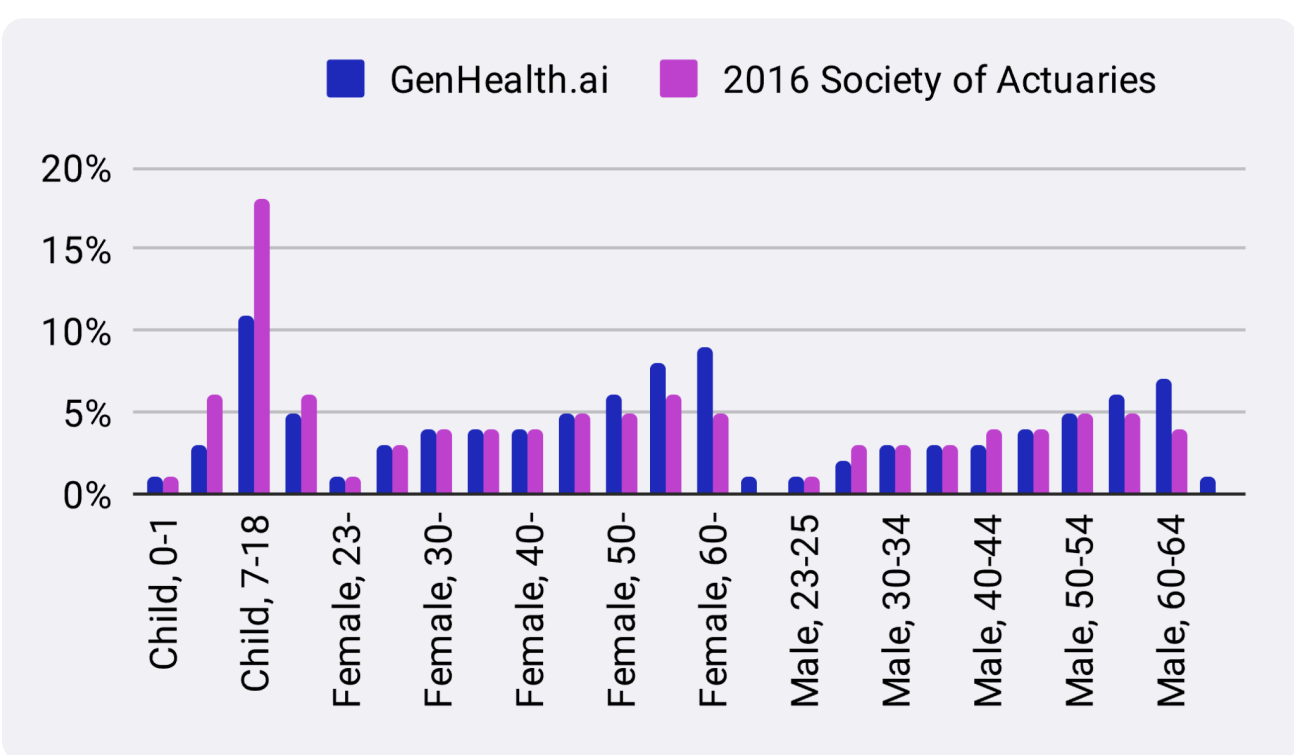

**Figure 2: Population Age and Gender Distributions**

Figure 2 shows the comparison between demographic distribution of the GenHealth cohort with the 2016 SoA study cohort. Our sample, derived from 2017 and 2018 commercial databases, closely aligns with the reference population of the 2016 SoA cohort, which was sampled from the 2012 and 2013 MarketScan databases. The observed differences reflect a slightly aging population.

#### 4.1.3 Measure of fit

This section introduces the key metrics used in the SoA's research to evaluate the accuracy and effectiveness of the LMM in estimating healthcare costs.

**R-Squared**

R-Squared ($R^2$, Coefficient of Determination) measures the proportion of the variance in the actual costs that is predictable from the predicted costs. It is calculated as:

$$R^2 = 1 - \frac{\sum_{i=1}^{n}(y_i - \hat{y}_i)^2}{\sum_{i=1}^{n}(y_i - \bar{y})^2}$$

where $\hat{y}_i$ is predicted cost, $y_i$ is actual cost, $\bar{y}$ is the mean of the actual costs, and $n$ is the number of observations.

A higher $R^2$ value suggests that the model more effectively captures the underlying patterns in the data, leading to more accurate predictions.

**Normalized Mean Absolute Error (NMAE)**

Mean Absolute Error (MAE) measures the average magnitude of the errors between predicted and actual costs, without considering their direction. It is defined as:

$$MAE = \frac{1}{n}\sum_{i=1}^{n}|\hat{y}_i - y_i|$$

By normalizing the MAE, we account for variations in scale, allowing for a more meaningful comparison of model performance across different datasets and contexts. A lower NMAE indicates better model accuracy, with values closer to zero reflecting predictions that are more precise relative to the scale of the data.

#### 4.1.4 Results

**The LMM reduces the Normalized Mean Absolute Error by approximately 14.1% compared to the best commercial models.** The SoA conducted analysis on both a raw uncensored population and a censored population (where patients were removed from the metric calculations if their predicted cost was greater than $250,000). As shown in Table 1, we find that our model outperforms on both uncensored and censored calculations, and for the purposes of this research, we report the data for the raw uncensored comparisons. The LMM achieves a NMAE of 78.3%, which is 14.1% better than the previous best NMAE of 91.2% reported in the 2016 SoA study. The average cost in the GenHealth cohort is $6,097 so on average the LMM would predict $972.closer to actual cost compared to the best model from the SoA paper. To calculate the future year cost, we run the Monte Carlo simulation on 64 generated futures for each patient and sum the dollar amount tokens from each future to create that simulation's predicted cost. Finally, we average the 64 futures' total costs to produce the model's predicted cost for each patient.

| | R-Squared (higher=better) | NMAE (lower=better) |
|---|---|---|
| Cotiviti (DxCG) | 23.8% | 91.2% |
| Hopkins (ACG) | 17.8% | 96.7% |
| Milliman (MARA) | 24.8% | 91.8% |
| SCIO (PCCM) | 15.1% | 95.8% |
| US Gov (HHS-HCC) | -34.3% | 122.8% |
| **GenHealth.ai (LMM)** | **25.3%** | **78.3%** |

**Table 1: R-Squared and MAE comparison with other models**

The model also achieves an $R^2$ (Coefficient of Determination) of 25.3%, which is a 2% improvement over the best model in the SoA study. This suggests that in addition to predicting better cost on average, the model's explanatory power of the variance is also improved. When compared with the HHS-HCC model ("The HHS-HCC Risk Adjustment Model for Individual and Small Group Markets under the Affordable Care Act - Centers for Medicare & Medicaid Services", n.d.) — a risk adjustment methodology commonly used in the US healthcare industry—on the same cohort, it produced a NMAE of 123% and a negative $R^2$ of 0.343, indicating a poor fit, whereas our model demonstrates better explanatory power and overall performance. For reference, the square of the Pearson correlation coefficient ($r^2$) for HHS-HCC is 0.134 vs 0.282 for the GenHelath.ai LMM.

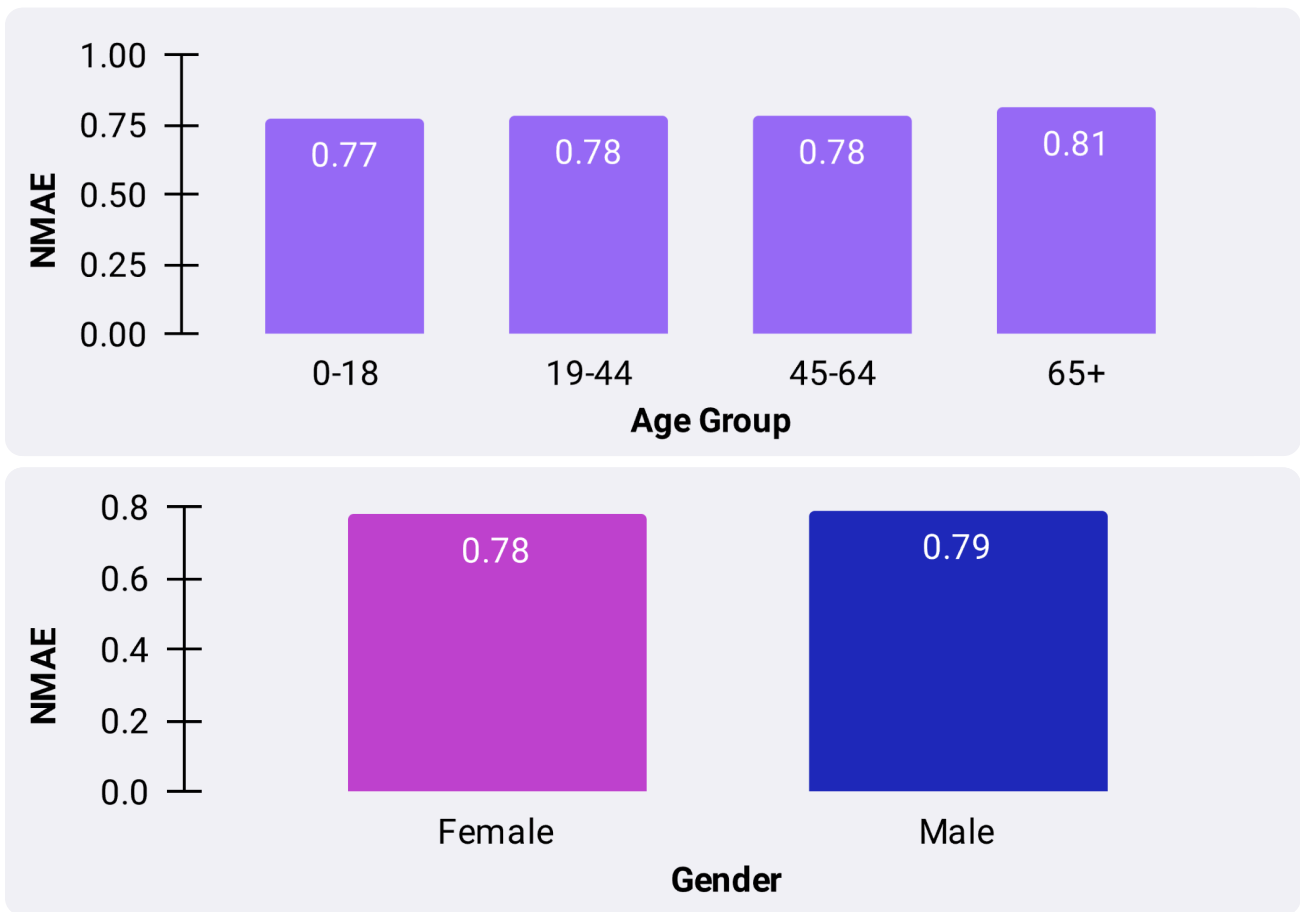

**Figure 3: NMAE Across Demographic Categories**

To address potential biases related to gender and age in the LMM, we analyzed the data across various demographic categories as shown in Figure 3. Overall, no significant biases were detected, with NMAE values across age groups ranging between 0.77 and 0.81, and across gender, between 0.78 for females and 0.79 for males, suggesting that the model does not exhibit significant bias toward any particular group.

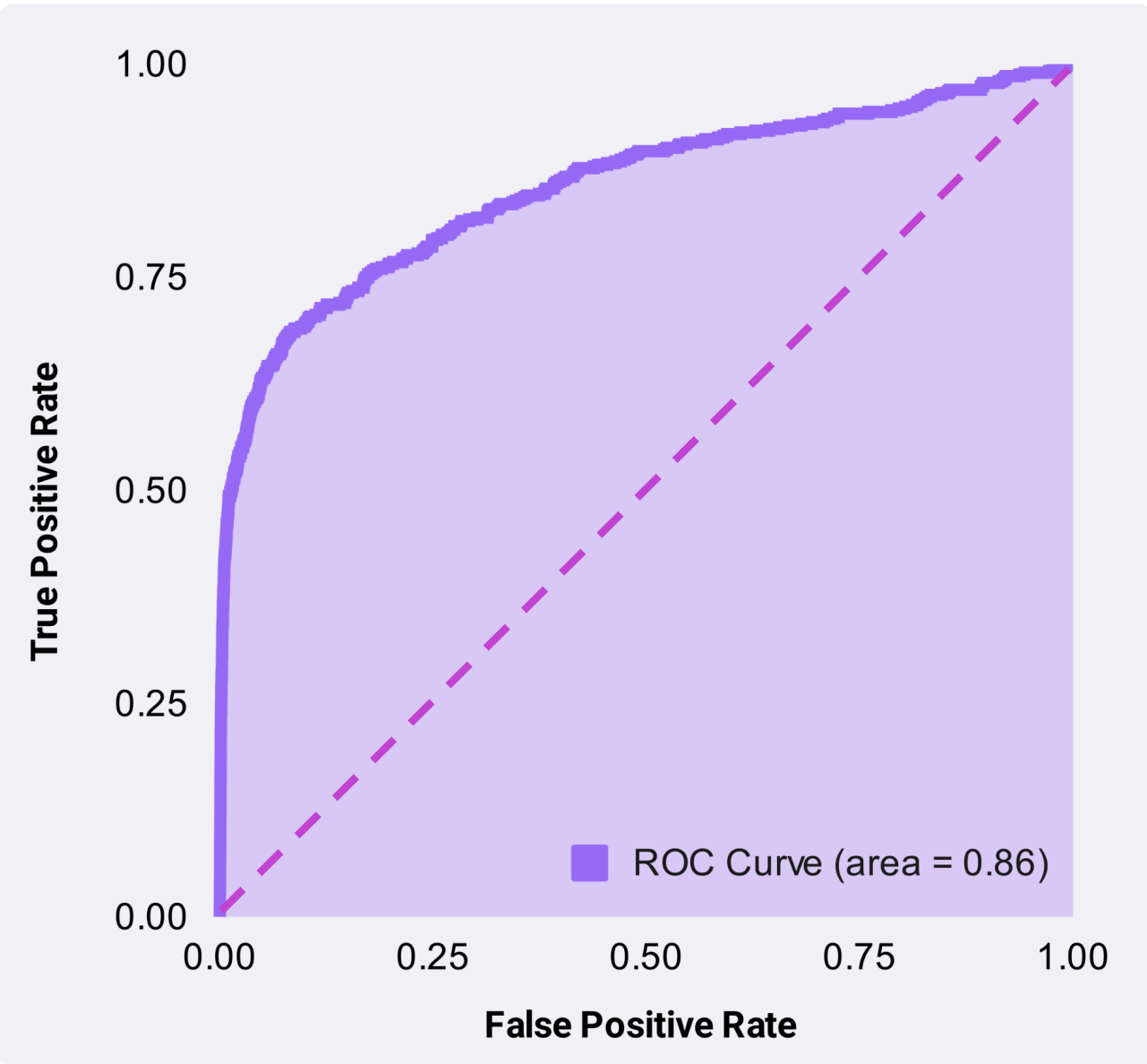

**Figure 4: Top 1 Percent Identification - AUC**

The analysis of cost percentiles shows a strong alignment between predicted and actual costs in the lower and mid-range percentiles with greater divergence in the highest percentile. This error in the tail end of heavily skewed cost distribution of healthcare is to be expected, yet an analysis of identifying the top 1 percent of high cost individuals yields an AUC score of 0.86 (on par with the best models in SoA) as can be seen in Figure 4.

The LMM shows improved accuracy and consistent performance in predicting healthcare costs. It improves upon existing benchmarks while maintaining equity across demographic groups and provides an option to view distributional output, not just point estimates.

As a by-product of the cost data, the LMM also produces events for predicted conditions, procedures, drugs, and more over time. Next, this study also evaluates the efficacy of the conditions, and more specifically chronic disease prediction.

## 4.2 Chronic Disease Prediction

Unlike the risk models in the SoA paper, the LMM has the ability to predict any condition, procedure, medication, or episode of care etc. and the time to those events. Chronic condition prediction is often critical in managing care. To assess the LMM's ability to predict diagnoses we utilized definitions of chronic conditions from the Chronic Conditions Data Warehouse (CCW) for evaluating predictions across 30 chronic diseases categories and compared our results to *BERHT* introduced a novel approach to modeling longitudinal patient data using transformers within electronic health records (EHRs) for predictive healthcare tasks. While the LMM predicts ICD diagnosis codes directly, BERHT predicts a higher level aggregation for 301 conditions. On aggregating across common conditions we are able to directly compare 19 chronic condition categories with BERHT and find that the LMM has marginally better results.

### 4.2.1 Data Collection

In this CCW section, we leveraged data from both commercial insurers and the Centers for Medicare & Medicaid Services (CMS), and randomly selected 50,000 records from each source's validation set for evaluating performance. Our focus was on patients who were enrolled in 2017 and continued their enrollment through 2021, allowing us to capture a comprehensive longitudinal view of their health trajectories. To ensure consistency with the BEHRT study, which emphasizes predictions within the first six months of the year, our analysis was concentrated on this same prediction period.

Note the difference in data selection from the SoA paper. We included the Medicare population (over 65 of age) and we included up to 4 years of history instead of only 1 year.

In order to compare our results with those in BEHRT, we mapped 19 conditions from the CCW dataset to their equivalents in the Caliber codes which the BEHRT paper used. For example, we mapped "Alzheimer's disease" and "non Alzheimer's dementia" from CCW both to Dementia in the Caliber framework. Through this mapping process, we achieved a total of 19 conditions that were consistently evaluated across both datasets, enabling a comparison of chronic disease prediction outcomes. The full mapping is in Appendix A.

### 4.2.2 Data Overview

The Figure 6 displays demographic distributions for the evaluation cohort including age, gender, and ethnicity. The age distribution covers a range from 0 to over 100 and 7 ethnicities. The lower count in the 60-70 group is due to US patients transitioning to Medicare Advantage resulting in fewer patients whose histories include 4 years which we maintained for consistency to BEHRT.

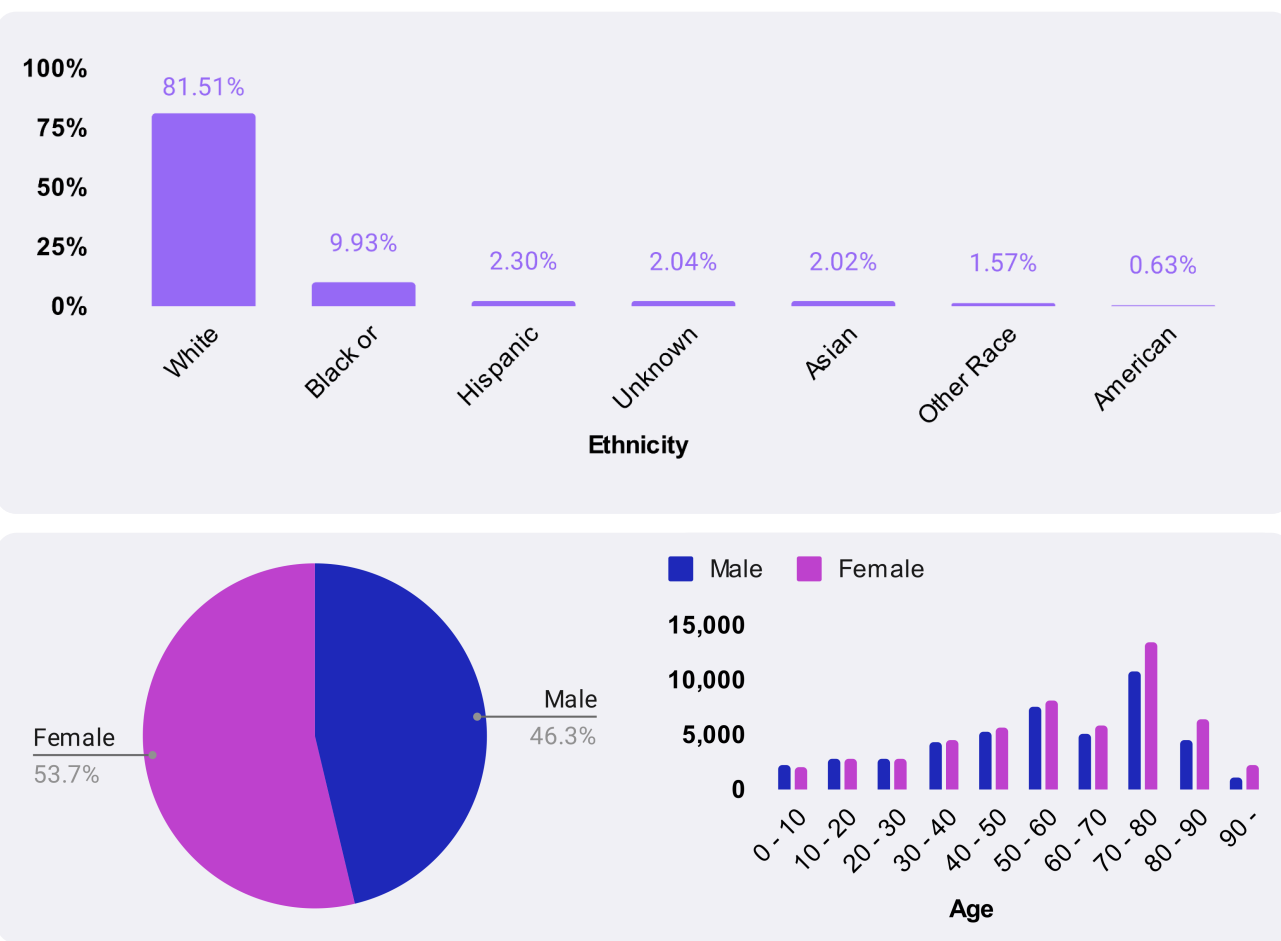

**Figure 6: Demographic Age, Gender, and Ethnicity Distribution**

Due to the cohort criteria requiring patients to be enrolled for at least four years and separate identifiers between our commercial and Medicare datasets there are fewer individuals in their mid-60's as individuals transition plan types and lack the full four year continuous history required for the cohort.

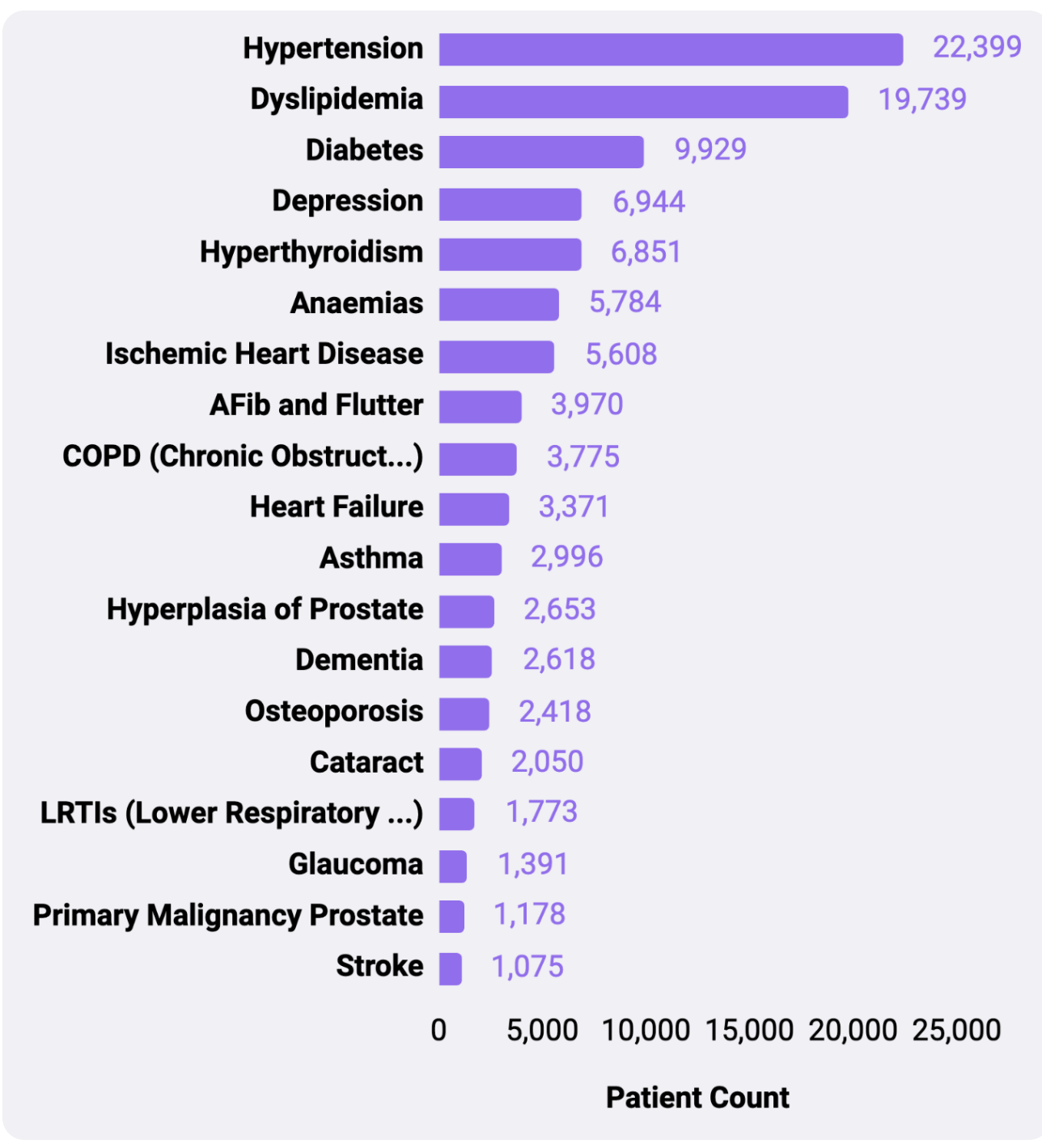

**Figure 7: Number of patients by Disease**

Figure 7 displays the total number of cases per disease, highlighting the distribution of patients across various conditions in the dataset. Conditions like hypertension and hyperlipidemia have the highest occurrence ratios, with significantly more patients compared to rarer conditions such as endometrial cancer and hip pelvic fracture.

### 4.2.3 Measures of fit

To compare our model with the BEHRT model we will compare AUROC and AUPRC scores.

**AUROC (Area Under the Receiver Operating Characteristic Curve):** This metric measures the model's ability to distinguish between classes at all decision thresholds. The area under the curve reflects the probability that the model will rank a randomly chosen positive instance higher than a randomly chosen negative one.

$$AUROC = \int_0^1 TPR(t)\, dFPR(t)$$

where:

- **True Positive Rate (TPR)** or **Sensitivity** is defined as:

$$TPR = \frac{TP}{TP + FN}$$

- **False Positive Rate (FPR)** is defined as:

$$FPR = \frac{FP}{FP + TN}$$

**APS (Average precision Score) or AUPRC (Area Under the Precision-Recall Curve):** This metric emphasizes the trade-off between precision and recall across various thresholds, making it particularly valuable for evaluating performance on imbalanced datasets.

$$AUPRC = \int_0^1 Precision(r)\, dRecall(r)$$

where:

- **Precision** is defined as:

$$Precision = \frac{TP}{TP + FP}$$

- **Recall** (same as TPR) is defined as:

$$Recall = \frac{TP}{TP + FN}$$

It should be noted that in cases of class imbalance, like rare disease prediction data, APS offers a more nuanced view of model performance. However, this metric is influenced by the number of positive cases in the dataset. In other words, as the prevalence of the positive class decreases, the number of false positives is likely to increase because the model encounters more negative instances that could be incorrectly classified as positive. This decrease in precision can significantly affect the shape of the precision-recall curve and therefore the AUPRC. Because of this it can be difficult to compare APS across evaluation sets with significantly different baseline prevalence rates.

In contrast, while AUROC provides a general measure of a model's ability to distinguish between classes, it may overestimate performance in datasets with a low prevalence of the positive class. This is because the true positive rate can heavily influence overall results, even if the model struggles with detecting rare positive cases.

Using both AUROC and AUPRC together can give a clearer picture of model performance, especially for rare disease predictions where accuracy in detecting the minority class is critical.

### 4.2.4 Results

Across the 19 conditions mapped using CCW and Caliber Codes, the GenHealth LMM achieves an average AUROC of 0.897 which is a 1.9% improvement over the BEHRT score of 0.881. We see close performance between the two models, with minor variations by conditions. The LMM sequences, however, indicate all other events before and after the chronic condition diagnosis; that level of detail in output can make the LMM more actionable.

Specifically, in Figure 8, the GenHealth model demonstrates higher AUROC scores for conditions like diabetes (0.95 vs. 0.81), atrial fibrillation and flutter (0.95 vs. 0.90), dyslipidemia (0.87 vs. 0.79), and hypertension (0.90 vs. 0.82).

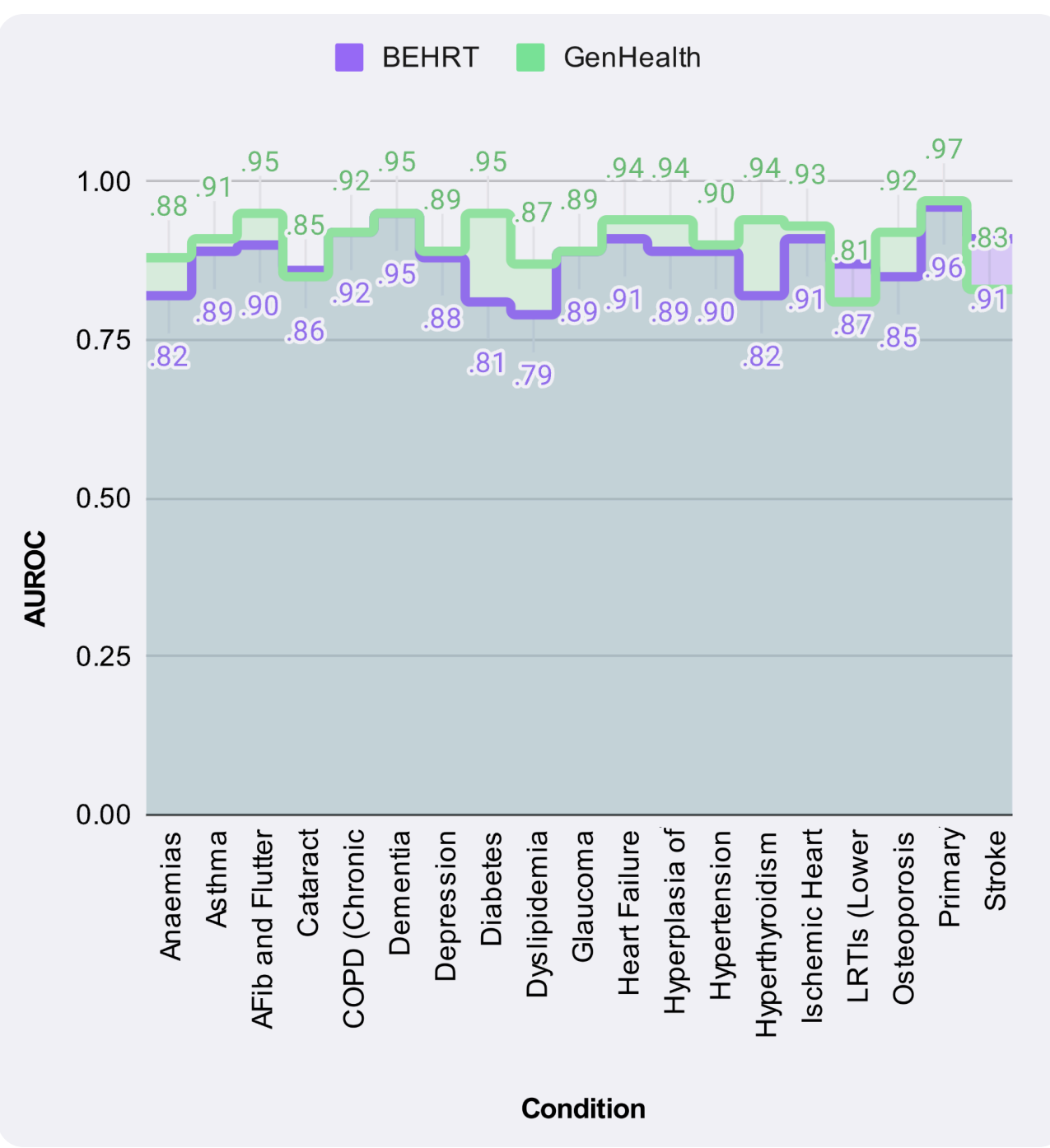

**Figure 8: Genhealth AUROC compared with BEHRT'sAUROC**

In contrast, the BEHRT model has higher AUROC scores for conditions such as stroke (0.91 vs. 0.83) and lower respiratory tract infections (0.83 vs. 0.91).

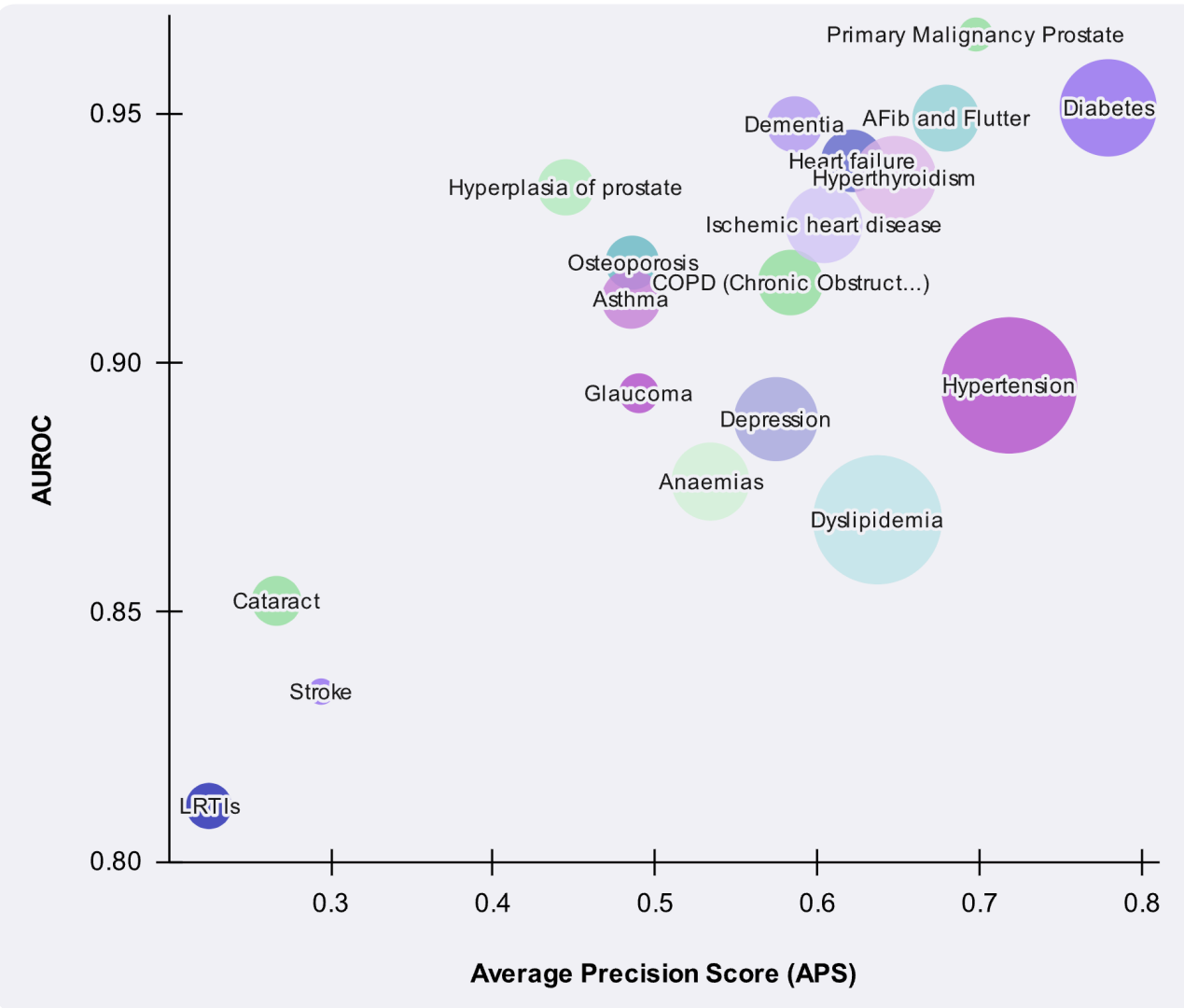

**Figure 9: Average Precision and ROC score for Disease Prediction**

Figure 9 illustrates the performance of the LMM across the 19 final diseases, represented by two key metrics: the APS on the x-axis and the ROC Score on the y-axis. Each point corresponds to one disease, with the size and color of the points reflecting the Occurrence Ratio (Occurence / Patient population) of the disease in the dataset.

Notably, diabetes, hypertension, and dyslipidemia are diseases where the model demonstrates both high APS and ROC Scores. The high APS for these conditions is expected given their higher prevalence, which typically makes it easier to achieve strong precision-recall performance. Moreover, the high AUROC scores for these more common diseases indicate that the model's ability to discriminate between classes remains robust even in more balanced datasets. This combination of high APS and high AUROC attests to the model's accuracy and reliability in predicting these more common diseases.

However, it's important to note that other conditions with lower prevalence tend to have higher AUROC scores. This is because AUROC is less effective in imbalanced datasets, often resulting in artificially inflated scores for conditions with lower prevalence.

For diseases with relatively lower prevalence but still high APS and ROC Scores, atrial fibrillation and flutter, hypothyroidism, heart failure and primary malignancy prostate stand out. Despite their lower occurrence, the model maintains a strong ability to discriminate between classes and balances precision with recall effectively. This indicates that the model is well-calibrated not only for prevalent conditions but also for certain less common, yet clinically significant diseases.

By considering both metrics, particularly in datasets with varying prevalence rates across different diseases, we ensure a comprehensive and accurate assessment of the model's effectiveness across a diverse range of conditions.

## 5 Discussion and Future Research

Our research introduces the Large Medical Model (LMM), which, for the first time, combines numerous advancements building over prior models: 1) we sequenced **medical event tokens** rather than natural language text or one-hot-encoding 2) we employed a **transformer** based architecture 3) we achieve a **large scale** by training on 140M patients 4) we **simulated multiple futures** using a Monte Carlo method to predict any of those events including cost. Our results suggest that the model is capable of understanding and modeling a varied set of patient timelines and is able to outperform top risk scoring models by 14.1% on cost of care and 1.9% on disease risk prediction over a range of conditions.

The LMM marks a significant advancement in the application of transformer-based architectures over previous state of the art. In particular, we demonstrate the model's effectiveness in key downstream tasks such as cost and chronic disease prediction, finding it to perform exceptionally well despite only being (pre)-trained on next token prediction. In doing so, we set a new benchmark for the integration of such models in real world health care settings, enabling more precise and personalized healthcare interventions.

Most notable, however, is the unique method of sequencing medical events over time. By training the model on temporally accurate sequences of actual patient histories, the LMM can generate more types of actionable insights than any existing risk and cost prediction methods. For example, the LMM can not only stratify a population based on cost or risk for some condition, but it can also identify specifically why the most complex and expensive patients in that group will be expensive. The LMM can specify the individual procedure codes and conditions that the patient may have in the future and when they may occur. Additionally, it can also identify the likelihood of a patient to be admitted to an Emergency Department, what that admission is expected to be for, and how much that will cost. This level of detail is unprecedented in the industry today and is crucial for positively impacting patient care and healthcare costs.

## 5.1 In-Silico Research and Personalized Medicine

In addition to its ability to predict total cost of care and onset of conditions, the LMM is also able to simulate the clinical effects of an intervention, new diagnosis, and more. This can be achieved by appending an additional event to the end of any patient history used as input into the model. We explored some of these simulated interventions using the LMM and see that it can identify novel clinical relationships unexplored in the current medical literature. For example emerging but inconclusive research exists investigating hearing stroke as a cause of Parkinson's. Using the LMM, we can simulate the effect of stroke on various patients including a 70 year old female and males. The LMM indicates a much higher likelihood of the patient developing Parkinson's and Parkinsonism for males than females. Although our model has never before seen any research papers or text indicating causality (given that it is trained only on medical event codes), our findings are corroborated by published research by the Parkinson's Foundation in which men have 1.5 times higher likelihood of Parkinson's ("Prevalence of Parkinson's disease across North America"). That research however does not indicate what triggers the diagnosis. The LMM, however, indicates it is the stroke that leads to Parkinson's. This relationship has been found in mice models ("Ischemic stroke causes Parkinson's disease-like pathology and symptoms in transgenic mice overexpressing alpha-synuclein") and in associations without direct casualty ascribed in human data research ("Associations between cerebrovascular risk factors and parkinson disease").

See the higher likelihood of Parkinson's-related events (colored in black) after a 70-year-old has a stroke. (Figure 10).

**Females** **Males**

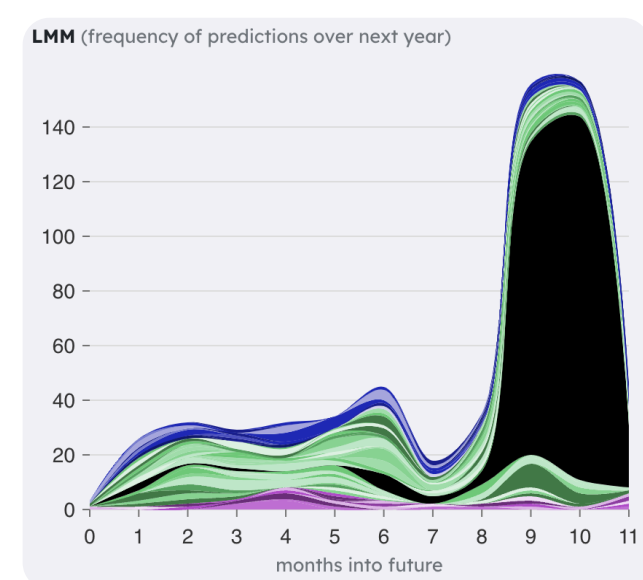

**Figure 10: Simulated likelihood of Parkinson's and Parkinsonism events after a stroke for 70-year-olds, with Parkinsonism events colored in black.**

Unlike traditional research, the Large Medical Model is capable of running in-silico in minutes without the complexity of live subjects or data collection for relevant cohorts. If each patient's record were armed with this information of potential future events, we imagine treatment of care would be far better informed. Ultimately, we expect use of the LMM to provide an efficient means of exploring medical relationships more broadly and to its ability to aid the selection of in-vivo studies.

## 5.2 Use Cases

Like the above, the LMM's predictions can be leveraged to address a broad set of healthcare challenges beyond the evaluations detailed in this paper. Similar to LLMs which can be used to perform Q&A, run chatbots, write articles, summarize text, etc., the LMM supports a wide range of use cases that are crucial for the healthcare industry. For instance, in **Population Health**, the LMM can stratify patient populations and identify optimal next-best actions by analyzing care paths derived from over 100 million other patients. For **Prior Authorization**, the model can simulate patient journeys to better understand outcomes and cost implications, thereby improving decision-making during care determinations. **Financial Forecasting** is another key application, where the LMM can predict the total cost of care using simulated care pathways for individuals or cohorts, enhancing underwriting, stop-loss insurance, and financial planning. Lastly, in **Risk Management**, the LMM can forecast patient health trajectories at the ICD level, enabling the identification of high utilizers and improving care paths.

In future research, we will explore a broader range of downstream tasks to demonstrate additional domains where this approach can positively impact healthcare.

The Large Medical Model represents a leap forward in how we can model and understand healthcare, leveraging the advances in modern AI and a deep understanding of the healthcare domain. We hope the broader use of it in the industry will lead to a more efficient healthcare system which can also significantly improve patient care.

**Acknowledgments**

Special thanks to Geof Hileman FSA, MAAA (author of the referenced Society of Actuaries paper) for reviewing this manuscript.

**Appendix A**

Mapping conditions between CCW and Caliber Conditions from 57 reported disease in Table 5 of BEHRT paper indicating where there are minor variations:

| CCW Condition | Mapped Caliber Condition |
|---|---|
| Acute myocardial infarction | (not in BEHRT) |
| Anemia | Other anaemias + Iron deficiency anemia + Vitamin B12 deficiency anemia |
| Asthma | Asthma |
| Atrial fibrillation and flutter | Atrial Fibrillation and flutter |
| Benign prostatic hyperplasia | Hyperplasia of prostate |
| Cancer breast | (not in BEHRT) |
| Cancer colorectal | (not in BEHRT) |
| Cancer endometrial | (not in BEHRT) |
| Cancer lung | (not in BEHRT) |
| Cancer prostate | Primary Malignancy Prostate |
| Cancer urologic | (not in BEHRT) |
| Cataract | Cataract |
| Chronic kidney disease | (not in BEHRT) |
| Chronic obstructive pulmonary disease | Chronic obstructive pulmonary disease (COPD) |
| Dementia (Alzheimer's disease + non-Alzheimer's dementia) | Dementia |
| Depressive mood disorders | Depression |
| Diabetes | Type 1 Diabetes Mellitus, Type 2 Diabetes Mellitus, and Diabetes Mellitus – other or not specified |

| Glaucoma | Glaucoma |
|---|---|
| Heart failure<br>non-ischemic heart disease | Heart failure |
| Hip pelvic fracture | (not in BEHRT) |
| Hyperlipidemia | Dyslipidemia |
| Hypertension | Hypertension |
| Hypothyroidism | Hypo or hyperthyroidism |
| Ischemic heart disease | Coronary heart disease not otherwise specified |
| Osteoporosis | Osteoporosis |
| Parkinsons | (not in BEHRT) |
| Pneumonia | Lower Respiratory Tract Infections |
| Rheumatoid arthritis | (not in BEHRT) |
| Stroke | Stroke Not otherwise specified (NOS) |